# TabPFN for Zero-shot Parametric Engineering Design Generation

Ke Wang[1], Yifan Tang[1*], Nguyen Gia Hien Vu[1], Faez Ahmed[2], and G. Gary Wang[1]

**Abstract**

*Deep generative models for engineering design often require substantial computational cost, large training datasets, and extensive retraining when design requirements or datasets change, limiting their applicability in real-world engineering design workflow. In this work, we propose a zero-shot generation framework for parametric engineering design based on TabPFN, enabling conditional design generation using only a limited number of reference samples and without any task-specific model training or fine-tuning. The proposed method generates design parameters sequentially conditioned on target performance indicators, providing a flexible alternative to conventional generative models. The effectiveness of the proposed approach is evaluated on three engineering design datasets, i.e., ship hull design, BlendedNet aircraft, and UIUC airfoil. Experimental results demonstrate that the proposed method achieves competitive diversity across highly structured parametric design spaces, remains robust to variations in sampling resolution and parameter dimensionality of geometry generation, and achieves a low performance error (e.g., less than 2% in generated ship hull designs' performance). Compared with diffusion-based generative models, the proposed framework significantly reduces computational overhead and data requirements while preserving reliable generation performance. These results highlight the potential of zero-shot, data-efficient generation as a practical and efficient tool for engineering design, enabling rapid deployment, flexible adaptation to new design settings, and ease of integration into real-world engineering workflows.*

## 1. Introduction

Due to the inherent high complexity and significant impact on life-cycle costs, engineering design is a crucial task that requires careful planning and execution [1]. The success of a product design, evaluated from technology, economy, and ecology perspectives, remains constrained by designers' experience, knowledge, and skills [1]. To foster creativity, designers employ a variety of systematic methods, such as brainstorming [2], mind map [3], checklist [4], lateral thinking [5] to merge the group's ideas and explore more solutions. In addition to these creativity techniques, Parametric Design (PD) is another paradigm to constrain the design features by parameters and rules, enabling the designer to find new designs via parameter manipulation [6]. Alongside conceptual design exploration, the advancement of digital technology allows engineers to estimate the performance of products under development through simulations provided by various Computer-Aided Engineering (CAE) and Computer-Aided Design (CAD) software, thereby reducing the need for costly prototyping and physical experiments [7]. Building upon PD and CAE, the design parameters could be optimized to find a high-performance solution [8]. However, common CAE tools such as computational fluid dynamics (CFD) and finite element analysis (FEA) solvers are computationally expensive, especially for high-fidelity simulations [9]. As a result, the conventional engineering design pipeline struggles to efficiently explore novel design spaces and accelerate design cycles.

Advanced machine learning methods address these challenges by enabling rapid performance prediction [10] and data-driven generation of novel design candidates [11]. Among these approaches, generative models employ deep networks to generate new designs based on information learned from the input design dataset [12], providing a powerful tool for generative design via interpreting designer-specified requirements to automatically generate design candidates that satisfy prescribed functional constraints [13]. The mainstream generative models, Variational Autoencoders (VAE) [14], Generative Adversarial Networks (GAN) [15], and Denoising Diffusion Probabilistic Models (DDPM) [16], have been applied to the design generation of ship hulls [17,18], cement microstructure [19], wheel topologies [20], metamaterials [21], and components design [22] or CAD model simplification [23]. Furthermore, coupling generative models with conventional optimization algorithms enables efficient and accelerated solutions for topology optimization and engineering design optimization [20,24–27]. Among all those generative models, diffusion models outperform GAN-based and flow-based generative models in both sample quality and likelihood metrics [28]. Considering extra generation condition based on the diffusion models' inference process, the conditional diffusion

[1] Product Design and Optimization Laboratory, School of Mechatronic Systems Engineering, Simon Fraser University, Surrey, BC, V3T 0A3. * Corresponding author: yta88@sfu.ca

[2] Department of Mechanical Engineering, Massachusetts Institute of Technology, 77 Massachusetts Avenue, Cambridge, Massachusetts 02139

model (cDDPM) can further control output data [16] and show outstanding performance in design generation [18] and topology optimization [24]. Therefore, only diffusion models are compared with the proposed method in this work.

However, diffusion models and other generative models mentioned above have a common limitation, i.e., the limited generalization capability across generation tasks and conditions. A well-trained generative model struggles to be implemented in unseen tasks and conditions, e.g., using a ship hull generator in car design generation, or introducing new performance indicators as generation conditions. Therefore, in engineering design practice, changes of generation objects or expected performance indicators require the retraining of the model, resulting in extra computational cost.

To overcome these challenges, in-context learning provides a compelling alternative by allowing models to perform inference on new tasks using only a limited number of reference samples without additional training. In this context, TabPFN [29], i.e., Prior-Data Fitted Network (PFN) [30] for tabular data, have shown a promising generalization capability across engineering prediction tasks, such as validity and safety prediction, among eight engineering design datasets, including Airfoil, FRAMED, Solar, Three-bar Truss, and Welded Beam [31]. The pre-trained TabPFN model can be adapted to different prediction tasks by loading different reference input-output pairs [29], indicating that it can be applied to different tasks without retraining. However, TabPFN can only provide reliable results for one-dimensional prediction, whereas engineering design typically involves multiple design parameters. Ma et al. [32] proposed TabPFNGen, which uses TabPFN to build an energy-based generative model but cannot perform conditional generation. Therefore, we propose a new method that applies TabPFN to generate parametric designs with expected performance in a sequential manner, inspired by the Recurrent Neural Network (RNN) [33]. Through this method, we achieve three contributions:

1. A zero-shot conditional generative method is proposed for engineering design with a small reference design set. According to the authors' knowledge, this is the first generative framework with TabPFN for conditional engineering design.
2. The proposed method allows generalization to diverse parametric design datasets without retraining.
3. The proposed method supports partial design completion and flexible conditioning (e.g., inferring the design parameters of the stern of a ship based on other known design parameters and expected performance).

In this work, we evaluate the performance of the proposed method across different engineering design problems, i.e., Ship hull [17,34], BlendedNet [35], and UIUC airfoil [36]. All training and inference processes are operated on the platform with RTX6000 Pro Blackwell GPU.

The remainder of the papers are structured as follows. Section 2 reviews related works, with a focus on tabular prior-data fitted networks. Section 3 presents the proposed methodology, followed by evaluation metrics, the effect of reference data size, and the design inpainting strategy. Section 4 reports experimental results on multiple datasets, covering accuracy, diversity, computational cost, and design inpainting performance. Section 5 discusses the implications of the proposed method, its applicability to geometry generation, and its limitations. Finally, Section 6 concludes the paper.

## 2. Related Works
### 2.1 Tabular Prior-data Fitted Network

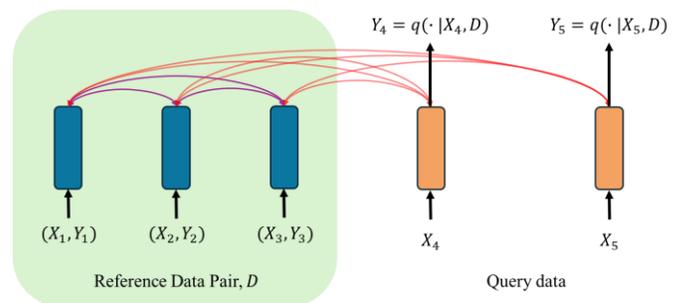

*Fig. 1 Inference procedure of TabPFN (revised from [29])*

The Tabular Prior-data Fitter Network (TabPFN) was proposed by Hollmann et al. [29] for prediction on small datasets by leveraging in-context learning within a Transformer architecture. Through learning from a large, complicated prior causal reasoning training set, TabPFN can do approximate Bayesian inference after offline training [29]. In the inference procedure of TabPFN as shown in Fig. 1, the pre-trained TabPFN model, $q$, should be fitted into a reference data pair $D$ consisting of multi-dimensional input $X$ and one-dimensional output $Y$, and then predict the output of the query input. For regression tasks, TabPFN regressor employs a piece-wise constant output distribution, which allows the model to predict a full probability distribution over target values rather than a single point estimate, and thereby represent uncertainty and multimodal outcomes [37]. The final regression output is the expected value of the predicted target distribution.

Through in-context learning capability, importing different reference data into a pre-trained TabPFN allows it to work in different tasks. For example, the pre-traiend TabPFN has already been applied to biogeographical ancestry predictions [38], pump optimization [39], machine fault classification [40], and disease diagnosis [41] for zero-shot application. Besides outstanding performance in classification and regression tasks, Ma et al. [42] used TabPFN as an energy-based model to generate tabular data with highly competitive results, but this method failed to provide a conditional generation solution. To address the above limitations, a sequential generation method with TabPFN is proposed for zero-shot engineering design generation with performance constraints.

## 3. Methodology

In this work, we propose a method that leverages the TabPFN regressor, which is provided by TabPFN v2 [37], to generate a new design in a sequential generation manner, as shown in Algorithm 1. In the proposed method, known designs and their corresponding engineering performance are first collected as reference data. At the initial step, the model is fitted to the reference set consisting of design performance and the first design parameter. The first parameter is then obtained solely from the target performance in the input query. Subsequently, the reference set is expanded to include the second design parameter, while the newly generated parameter is appended to the query input for the next generation step. This procedure is repeated sequentially until all design parameters are generated.

---

Algorithm 1 Parametric engineering design generation through sequential generation based on TabPFN

---

1. $C_{ref} = [P_1^{ref}, P_2^{ref}, ..., P_n^{ref}]$      # reference condition: performances of reference designs

2. $C_{gen} = [P_1^{gen}, P_2^{gen}, ..., P_n^{gen}]$      # generation condition: expected performances of generated designs

3. $X^{known} = [X_0^{known}, X_1^{known}, ..., X_N^{known}]$      # reference design with $N$ parameters

4. $Y_{gen} = [\,]$

5. For $i$ in $[0, 1, ..., N]$      # generate design parameters sequentially

6.      $X_{ref} = [C_{ref}]$ if $i = 0$ else $[C_{ref}, X_0^{known}, ..., X_{i-1}^{known}]$      # concatenate condition and known parameters to update the reference input

7.      $X_{gen} = [C_{gen}]$ if $i = 0$ else $[C_{gen}, Y_{gen}]$      # update query input for design parameter inference

8.      $Y_{ref} = X_i^{known}$      # update reference output data

9.      $TabPFN.fit([X_{ref}, Y_{ref}])$      # fit TabPFN to the reference set

10.      $Y_{gen} = [Y_{gen}, TabPFN(X_{gen})]$      # generate new data under the condition and concatenate it with the part generated before

11. Return $Y_{gen}$      # generation finish and return a complete design

---

In Algorithm 1, the first step is to collect all known designs and their performance as reference data, as shown in Line 1 and Line 3. The reference condition $C_{ref}$ refers to the known design's performances, and $P_i^{ref}$ is the $i$-th performance indicator of all reference designs (e.g., drag coefficient, and lift coefficient), and $n$ is the number of performance indicators of reference design. $X^{known}$ refers to designs used as reference with $N$ design parameters. In Line 2, the generation condition $C_{gen}$ include all expected performances of the generated designs, where $\{P_i^{gen}\}_{i=1}^{n}$ denote the expected values of the $n$ performance indicators. The number of the generated designs can be different from that of the reference design. Then, we need to build reference data consisting of input and output to fit the model before the generation. Therefore, in Lines 6 to 9, we update the reference input, $X_{ref}$, query input $X_{gen}$, and reference output $Y_{ref}$ according to the parameter we need to generate. The pre-trained TabPFN model is then fitted to the reference data, $[X_{ref}, Y_{ref}]$, to learn the internal relationship between the reference input and output. After that, we generate the new parameter and concatenate them together as the generated design $Y_{gen}$, according to the query data $X_{gen}$ (Line 10). Eventually, the generated design with complete design parameters is returned as the final output (Line 11). The workflow of the proposed method is presented in Fig. 2.

As shown in Fig. 2, at the initial stage of the proposed method, all target performances are used as conditions to generate the first design parameter. At each subsequent step, the newly generated parameter is concatenated with the existing conditions to construct the input for the next generation step. This sequential procedure is repeated until the complete set of design parameters is generated. In addition, this sequential generation method enables a flexible length in both generation conditions and generated design parameter. Through changing the composition of the generation condition, we flexibly import different numbers of performance constraints into the design generation and even introduce other non-performance elements (e.g. some known design parameters) into the generation condition to expand its application scenarios.

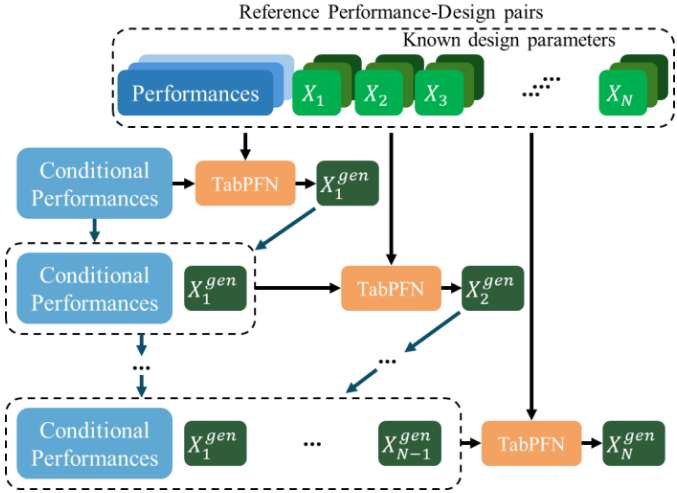

*Fig. 2 Sequential parametric design generation via TabPFN*

### 3.1 Evaluation Metrics

In this section, we design several comparison studies to evaluate the proposed method in terms of accuracy and diversity, on three engineering design datasets, i.e., Ship hull design, BlendedNet, and UIUC airfoil. Additionally, the conditional diffusion model (cDDPM), which has demonstrated outstanding performance in engineering design generation [17,18], is trained on the dataset mentioned above for comparison.

*3.1.1   Accuracy*

In this paper, we randomly pick 5,000 designs from the design data sets, using 70% of them as the reference set to fit the pre-trained TabPFN model and the remaining 30% as a test set for evaluating the design generation performance.

The mean absolute percentage error (MAPE) and mean absolute error (MAE) are calculated between the expected performances and the generated designs' performances. Instead of using expensive simulations, the performance of the generated design is obtained via formulation or TabPFN regression prediction. Besides that, we generated designs with the proposed method in two generation orders, the default parameter order provided by the design dataset and the random order, to determine the sensitivity of the model quality to the parameter generation order. In the latter case, We repeat the process of design selection, training–testing set splitting, and new design generation based on performance conditions from the testing set ten times, in order to compute the mean and standard deviation of both MAPE and MAE. The detailed results are discussed in Section 4.2.

*3.1.2   Diversity*

To assess the diversity of the generated designs and quantify how well the proposed method captures the variability of the target distribution, this work employs the Precision–Recall for Distributions (PRD) metric [43] together with the Maximum Mean Discrepancy (MMD) [44]. PRD provides a two-dimensional characterization of generative performance by separately quantifying precision and recall, enabling an informative evaluation of sample quality and sample diversity. The PRD between the generated distribution $Q$ and the reference distribution $P$ is computed following algorithm proposed by Sajjadi er al. [43]. Initially, both generated and reference distribution are defined on a finite state space, $\Omega$. A trade-off parameter $\lambda > 0$ is introduced to balance precision and recall. For each value of $\lambda$, the corresponding precision and recall can be calculated as:

$$\alpha(\lambda) = \sum_{\omega \in \Omega} \min(\lambda P(\omega), Q(\omega)) \quad (1)$$

$$\beta(\lambda) = \sum_{\omega \in \Omega} \min\left(P(\omega), \frac{Q(\omega)}{\lambda}\right) \quad (2)$$

The resulting pair $(\alpha(\lambda), \beta(\lambda))$ represents the maximal attainable precision and recall under the trade-off constraint $\alpha = \lambda \beta$, where $\lambda \in (0, \infty)$. To approximate the PRD curve, $\widehat{PRD}(Q, P)$, the pairs $(\alpha(\lambda), \beta(\lambda))$ are evaluated according to (1) and (2) over an equiangular grid of $\lambda$ values. Specifically, for a given angular resolution $m \in \mathbb{N}$, the PRD curve is computed as below.

$$\widehat{PRD}(Q, P) = \{(\alpha(\lambda), \beta(\lambda)) \mid \lambda \in \Lambda\}$$
$$\text{where } \Lambda = \left\{\tan\left(\frac{i}{m+1}\frac{\pi}{2}\right) \middle| i = 1,2,\dots,m\right\} \quad (3)$$

These points form an approximation of the PRD curve, which characterizes the full precision–recall trade-off between the generated and reference distributions. The curves that lie closer to the upper-right region indicate a more favorable balance between precision and recall.

Complementarily, MMD measures the discrepancy between the generated and reference distributions in a reproducing kernel Hilbert space. Thereby, MMD is sensitive to higher-order statistical differences and can capture global distributional alignment [45]. Given a set of reference samples $\{x_i\}_{i=1}^{n} \sim P$ and generated samples $\{y_j\}_{j=1}^{m} \sim Q$, MMD is defined as the squared distance between the mean embeddings of the two distributions in a reproducing kernel Hilbert space (RKHS).

$$MMD^2(P, Q) = \left\| E_{x \sim P}[\phi(x)] - E_{y \sim Q}[\phi(y)] \right\|_H^2 \quad (4)$$

where $\phi(\cdot)$ denotes the feature mapping induced by a positive-definite kernel $k(\cdot, \cdot)$. Then, MMD can be estimated from finite samples using the unbiased empirical estimator.

$$MMD^2 = \frac{1}{n(n-1)}\sum_{i \neq i'} k(x_i, x_i') + \frac{1}{m(m-1)}\sum_{i \neq i'} k(y_i, y_i') - \frac{2}{nm}\sum_{i \neq i'} k(x_i, y_i) \quad (5)$$

A Gaussian kernel is commonly adopted. A smaller MMD value indicates that the generated distribution $Q$ is closer to the reference distribution $P$. By jointly leveraging PRD and MMD, this evaluation framework offers a comprehensive assessment of generative diversity, balancing local sample fidelity with overall coverage of the target distribution. Results are given in Section 4.3.

*3.1.3    Data size and time cost*

The number of data points and time used for model training and design inference are compared to evaluate the efficiency of the proposed method and the compared method. Results are shown in Section 4.4.

### 3.2    Amount of reference data

After evaluating the proposed method's performance, we seek to identify the factors that affect it. The nature of the TabPFN is to predict new data based on relationships learned from the reference data [29]. The number of reference data determines the information the model can learn, influencing the final prediction performance. To determine the effect of the number of reference data, we sampled the number of selected designs as [200, 400, …, 10000] from the reference set where 10,000 is the maximum reference size of TabPFN. After that, 2,000 selected designs' performances are used as a condition to generate a new design and measure the performance error by the MAPE metric. The results are shown in Section 4.5.

### 3.3    Design Inpainting

Due to the sequential nature of generation, the proposed method can also perform design completion using partially known design parameter values. By concatenating the known design parameter values with the performance indicators to establish the generation condition, the proposed method can sequentially generate missing design parameters to achieve design inpainting or partial design regeneration tasks based on the other known design parameters. For example, in a ship hull design task, if several design parameters such as hull length and beam are already specified together with a target resistance value, the proposed method can treat these target values as conditions and sequentially generate the remaining unknown parameters (e.g., draft and stern shape). In this way, a complete and feasible hull design is obtained without altering the predefined parameters, enabling effective design inpainting and partial design regeneration. To assess the performance of the design inpainting task, we gradually increase the number of missing design parameters and generate new designs by completing the missing components to evaluate the accuracy of the proposed method under partial design completion, and place the results in Section 4.6.

### 4.    Results
### 4.1 Dataset and Preprocess

The UIUC airfoil database, provided by the University of Illinois at Urbana–Champaign [36], is a widely used benchmark dataset in aerodynamic shape analysis and airfoil generation tasks. It contains a large collection of two-dimensional airfoil geometries with diverse aerodynamic characteristics, including both classical airfoil families and custom-designed profiles. In this work, the original UIUC airfoil geometries are further preprocessed through uniform resampling to construct fixed-dimensional design representations suitable for generative modelling. Specifically, two resampled datasets with drag-lift ratio $C_l/C_d$ are created. In the first dataset, each airfoil surface is sampled with 30 points on the upper surface and 30 points on the lower surface, resulting in a total of 60 sampling points and a 120-dimensional design vector. In the second dataset, a higher-resolution representation is adopted, with 50 sampling points on each surface, leading to 100 sampling points and a corresponding 200-dimensional design vector. These two datasets enable a systematic evaluation of the proposed method under different geometric resolutions.

The ship hull design dataset comes from C-Shipgen [17] and ShipD [34], a parametric ship hull design dataset constructed for data-driven conceptual ship design. Each sample is represented by a fixed-dimensional design vector encoding the principal geometric characteristics of the hull, together with corresponding drag coefficient, $C_d$. Notably, the design space of ShipD [34] exhibits a hierarchical and mixed-type structure. Several Boolean variables are included to control the activation and interpretation of other continuous design parameters, such as the presence of specific hull features or the selection of different geometric formulations. In addition, partial continuous variables are defined as ratio-based parameters, whose values are expressed relative to other reference design variables rather than as absolute geometric quantities. This introduces strong conditional dependencies among parameters, resulting in a nontrivial and structured design space.

BlendedNet [35] is a publicly available high-fidelity aerodynamic dataset for blended wing body (BWB) aircraft, consisting of 999 parameterized geometries evaluated under multiple flight conditions using Reynolds-Averaged Navier–Stokes simulations. In addition to integrated aerodynamic coefficients, the dataset provides detailed pointwise surface pressure and skin friction distributions, enabling data-driven surrogate modeling at high spatial resolution [35]. In this work, we do not employ the full BlendedNet geometry reconstruction or surrogate modeling pipeline. Instead, we adopt only the ten independent geometric design parameters proposed in BlendedNet as the design representation and their drag coefficient $C_d$, lift coefficient $C_l$, and moment coefficient $C_{my}$ for our experiments. This allows us to focus on the generative modeling of structured and interpretable design parameters while avoiding additional complexity introduced by downstream geometric and aerodynamic prediction stages.

Since neither the UIUC airfoil dataset [36] nor the BlendedNet dataset [35] provides an explicit analytical or parametric formulation for directly computing performance metrics from the design parameters, a surrogate model is required to evaluate the generated designs. In this work, we employ the TabPFN-based regressor as a unified performance predictor for both datasets. The predictive accuracy of the adopted regressor is quantitatively assessed and reported in the Appendix.

## 4.2 Accuracy

Based on the processed data, four design generation tasks are conducted by the TabPFN with the proposed sequential generation method according to the method introduced in Section 3.1.1. In the case of ship hull design, we generate 2000 designs conditioned by randomly selected $C_d$ values across the whole dataset in both TabPFN and cDDPM. In addition, cDDPM models are also trained on the BlendedNet and airfoil datasets using their respective training sets, and new designs are generated based on performance values from the testing sets. Moreover, all designs generated by TabPFN follow the dataset's default ordering. For example, the generation of a ship hull design starts from the Length of the Bow Taper and ends in the Fillet Radius for the Stern Bulb according to the default order; the airfoil generation starts from the left vertex and generates one nearby point in each step. After that, the performance error of the generated designs is calculated based on the method introduced in Section 3.1.1 and shown in Table 1.

Table 1 Accuracy of generated designs from the proposed method, TabnPFN, and the comparison model, cDDPM.

|  | Ship hull | | BlendedNet | | | | | | Airfoil 50 | | Airfoil 30 | |
|---|---|---|---|---|---|---|---|---|---|---|---|---|
|  | $C_d$ | | $C_d$ | | $C_l$ | | $C_{my}$ | | $C_l/C_d$ | | $C_l/C_d$ | |
|  | MAE | MAPE | MAE | MAPE | MAE | MAPE | MAE | MAPE | MAE | MAPE | MAE | MAPE |
| TabPFN | **4.577E-5** | **1.223%** | 0.00732 | **5.679%** | 0.00732 | **4.960%** | 0.00346 | **6.230%** | **14.853** | **25.441%** | 6.029 | 18.183% |
| cDDPM | 4.875E-4 | 10.791% | **0.00637** | 7.638% | **0.00143** | 5.145% | **0.00325** | 8.133% | 67.780 | 88.343% | **3.605** | **6.797%** |

Table 1 compares the performance errors of the generated designs from the proposed TabPFN method and cDDPM across three engineering design datasets. In the case of ship hull design, the design set has the most complex geometry-control mechanism, including continuous and Boolean variables. The proposed method achieves outstanding results with an MAPE of only 1.223%, while the MAE is approximately 10 times lower than that of cDDPM. Furthermore, TabPFN's MAPE in BlendedNet is slightly lower across all three performance indicators, $C_d$, $C_l$, and $C_m$, while MAE values are larger, indicating that both cDDPM and the proposed method in the BlendedNet dataset are comparable. Additionally, with the change of the number of sample points in the airfoil design dataset, the performance of both the proposed method and cDDPM changes. On the airfoil with 30 sample points, cDDPM achieves better results, with lower MAPE and MAE, than the proposed method. However, with increasing the number of sample points to 50, cDDPM's result deteriorates immediately with 88.343% $C_l/C_d$ MAPE and 67.78 $C_l/C_d$ MAE, while the proposed method's MAPE only increases from 18.183% to 25.441%, and MAE increases from 6.029 to 14.853, indicating a more stable performance when the number of sample points increases. In summary, under the default parameter order, the proposed method can generate designs with higher accuracy with highly parametric design sets, such as the ship hull, and shows better stability in the geometry generation task with different numbers of samples.

Table 2 Performance error of random-order parameter generation

|  |  | MAE | | MAPE (%) | |
|---|---|---|---|---|---|
|  |  | Mean (*) | Std. | Mean (*) | Std. |
| Ship hull | $C_d$ | 7.438e-4 (↑) | 3.702e-4 | 15.477 (↑) | 6.709 |
| BlendedNet | $C_d$ | 7.068e-4 (↓) | 1.251e-3 | 6.020 (↑) | 1.418 |
|  | $C_l$ | 1.580e-3 (↓) | 1.450e-4 | 5.740 (↑) | 0.645 |
|  | $C_{my}$ | 3.289e-3 (↓) | 6.691e-4 | 7.487 (↑) | 1.203 |
| Airfoil 50 | $\frac{C_l}{C_d}$ | 2.415 (↓) | 0.872 | 9.089 (↓) | 5.171 |
| Airfoil 30 | $\frac{C_l}{C_d}$ | 2.425 (↓) | 0.826 | 10.726 (↓) | 6.586 |

Furthermore, we randomize the parameter generation order and repeat the design generation in all three datasets by 10 times. The results are presented in Table 2 with the mean and standard deviation (std.) values. The arrows in the bracket indicate the comparison with the results reported in Table 1, where (↑) denotes an increase and (↓) denotes a decrease. By comparing the results presented in Table 1 and Table 2, it is observed that changes in the generation order influence the generation performance differently across design scenarios. For the ship hull design dataset, altering the generation order leads to a substantial increase in both MAE and MAPE, indicating a degradation in generation accuracy, with a MAPE standard deviation of 6.709% suggesting a strong sensitivity to parameter ordering in this highly coupled design space.

In contrast, the BlendedNet dataset demonstrates a high level of robustness to order variation. The mean MAE of $C_l$ (0.00158) and $C_{my}$ (0.00325) remains similar to those reported in Table 2, ($C_l$: 0.00143, and $C_{my}$: 0.00325) except MAE of $C_d$ reduce form 0.00732 to 0.00074. Furthermore, the mean MAPE increases only marginally from around 5.6% (default order, $C_d$: 5.7%, $C_l$: 5.0%, and $C_{my}$: 6.2%) to stick around 6.5% (random order, $C_d$: 6.0%, $C_l$: 5.7%, and $C_{my}$: 7.5%). The mismatch between MAE and MAPE of $C_d$ could be the error concentrate on the low $C_d$ value area. These results indicate a reliable generation under different orders.

Moreover, the airfoil datasets exhibit a pronounced performance improvement when random-order generation is employed. In this case, the MAE decreases substantially from 14.853 and 6.029 to a mean value of 2.415 or 2.435 with a standard deviation of around 0.8, and the corresponding MAPE is reduced to approximately 10%, highlighting the effectiveness of order randomization for 2D geometric design tasks.

Overall, these results demonstrate that the proposed framework is particularly well-suited for design problems characterized by independent or weakly coupled parameters, as well as 2D geometric representations, where it achieves robust and even improved generation performance without requiring a fixed parameter ordering. For design problems involving more complex internal relationships among parameters, the proposed method is still capable of producing competitive results. However, its performance becomes more sensitive to the chosen generation order.

### 4.3 Diversity

According to Section 3.1.2, PRD and MMD of generated designs are computed, and the results are presented in Fig. 3 and Table 3.

As shown in Fig. 3, compared with cDDPM, the proposed method can generate designs with high precision but falls short in recall. This implies that the proposed method can learn the relationship between the performance indicators and design parameters to generate a design close to the reference design set, conditioned on new performance values, but with a weak coverage. This short coverage range is caused by the limited number of TabPFN input reference performance-design data. The limited reference data count (Maximum 10,000) restricts the amount of information the TabPFN can learn, making its generated designs hard to cover the entire design space.

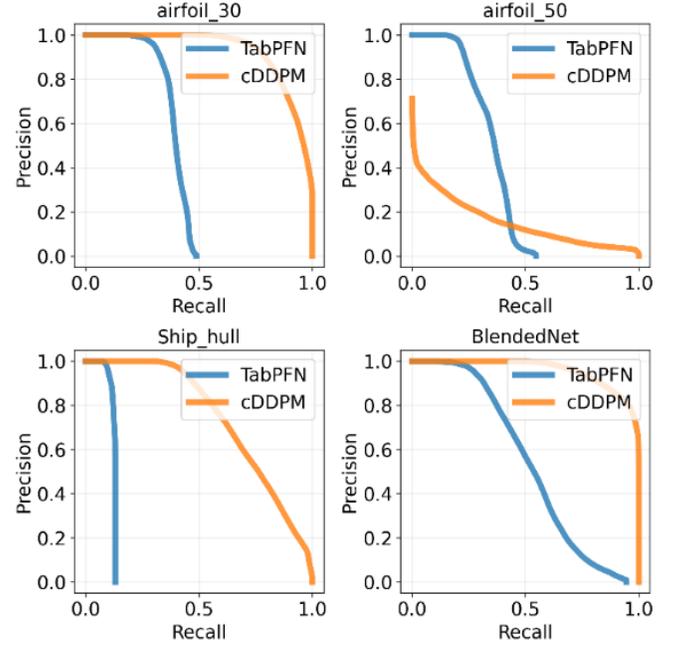

Fig. 3 Diversity of generated designs of the proposed method and the comparison model

In this work, the proposed method is expected to generate novel designs that are different from the training set but match the expected performance. The accuracy of the generated designs' performance was evaluated and discussed in Section 4.2, hence, in this part, MMD is used to evaluate how different the generated designs are from the training set. MMD measures the distance between two distributions, reflecting the similarity between the generated design distribution and distribution of the training data. A lower MMD value indicates a higher similarity between the two distributions, while a higher MMD value suggests a larger deviation. Consequently, MMD is used here as an indicator of design novelty, where lower MMD values correspond to lower novelty and higher MMD values indicate higher novelty.

Table 3 MMD values of the generated design of the proposed method and the comparison model

|  | Ship hull | BlendedNet | Airfoil 50 | Airfoil 30 |
| --- | --- | --- | --- | --- |
| TabPFN | 0.501 | 0.0472 | 0.0372 | 0.0196 |
| cDDPM | 0.0318 | 0.0558 | 0.0372 | 0.0372 |

From the MMD shown in Table *3*, the proposed method performs similarly to cDDPM for BlendedNet and airfoil with 50 sampling numbers, but worse for airfoil with 30 sampling numbers. However, the MMD of TabPFN's generated design (0.501) is significantly higher than that of cDDPM (0.0318).

This result shows that the proposed method can generate designs with similar or even greater novelty than using cDDPM in both the parametric design task (Ship Hull) and the high-sample-number geometry generation task (Airfoil 50). Furthermore, during experiments, authors observed that TabPFN exhibits strong consistency, meaning that the proposed method produces identical designs when the reference data and input conditions remain unchanged. To alleviate this behavior and promote greater output diversity under the same conditioning inputs, two strategies are proposed in the Appendix (Section A2), and their effectiveness is experimentally evaluated.

### 4.4 Data size and time cost

The data size and time used in the model training and designs synthesis by the method mentioned in Section 3.1.3 are presented in Table 4. As the inference time of proposed method increase with the number of reference data, the time cost of load maximum number of reference data (10,000 or the data number of training set) is measured and show in Table 4.

Table 4 Data size and time breakdown.

|  |  | cDDPM | TabPFN |
|---|---|---|---|
| Airfoil 30 | Training | 6h 25min | <1s (fitting) |
|  | Inference | 1.591s | <16min 12s |
|  | Data used | 38802 | <10000 |
| Airfoil 50 | Training | 8h 3min | <1s (fitting) |
|  | Inference | 2.175s | <44min 15s |
|  | Data used | 38802 | <10000 |
| Ship hull | Training | 7h 32min | <1s (fitting) |
|  | Inference | 1.700s | <6min 35s |
|  | Data used | 82168 | <10000 |
| BlendedNet | Training | 1h 38min | <1s (fitting) |
|  | Inference | 1.710s | <32s |
|  | Data used | 8830 | <8830 |

As shown in Table 4, the proposed TabPFN-based method requires a longer inference time compared to cDDPM across all evaluated datasets. However, this increased inference cost should be considered together with the substantially higher training cost required by cDDPM. In contrast, TabPFN operates in a training-free manner and only requires a lightweight fitting step, which consistently takes less than one second across all datasets. As a result, although the proposed method takes a higher inference time during design generation, it achieves a significantly lower overall computational cost when training time is taken into account. In addition, the proposed method requires considerably fewer reference samples (maximum 10,000) to generate new designs, which is markedly lower than the data volume required to train cDDPM models across the same datasets. These characteristics make the proposed method particularly attractive in data-limited or rapid-deployment scenarios. It is also observed that, as the dimensionality of the design space increases, the maximum inference time of the proposed method exhibits a nonlinear growth trend. This behavior reduces the relative time advantage of the proposed method for high-dimensional design generation tasks. A more detailed analysis of the scalability of the proposed method with respect to design dimensionality will be investigated in future work.

### 4.5 Effect of Reference Data Size

In this section, we further study the number of reference data to generate designs by the proposed method. According to Section 3.2, we randomly select 2000 ship hull designs from the ShipD dataset and then increase the number of reference designs from 200 to 9,800 by a step of 200 to generate new designs. The MAPE values of the drag coefficient $C_t$ is summarized in Fig. 4.

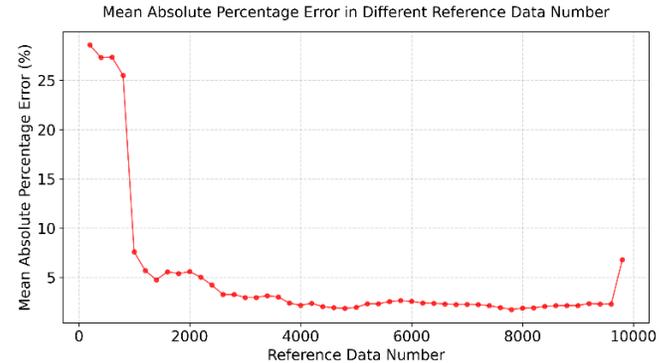

Fig. 4 MAPE of $C_w$ with different reference data sizes

As shown in Fig. 4, the MAPE of the generated ship hull designs exhibits a sharp decrease from 28.577% with 200 reference designs to 4.743% with 1400 reference designs. Following this point, a plateau emerges, with MAPE remaining relatively stable up to approximately 2000 reference designs. Beyond this region, the error continues to decline gradually, reaching 2.157% at 4000 reference designs, after which the curve displays only minor fluctuations. Notably, the MAPE shows a spike at 9800 reference designs but still stabilizes around 6.8%. Overall, increasing the number of reference designs consistently improves the accuracy of the generated designs. However, the marginal benefit diminishes substantially as the reference set grows.

### 4.6 Design Inpainting

The sequential generation method enables the proposed method to adapt to different numbers and types of conditions (e.g., performance indicators, environmental variables, and known design parameters). Therefore, users can use the proposed method to conduct a design inpainting task as introduced in Section 3.3. To evaluate the performance of the proposed method in the design inpainting task, we conduct an experiment based on the ship hull design data set following the method introduced in Section 3.3, and the results are presented in Fig. 5.

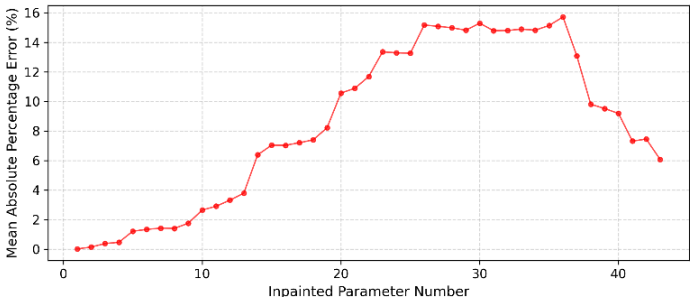

Fig. 5 Performance ($C_w$) MAPE of cases of different numbers of inpainted design parameters

As shown in Fig. 5, the mean absolute percentage error (MAPE) varies non-monotonically with respect to the number of inpainted design parameters. When only a small number of parameters are inpainted, the error remains low and increases gradually, indicating that the proposed method can reliably complete designs when sufficient contextual information is available. As the number of inpainted parameters increases to over 23, the MAPE rises sharply and reaches a plateau, suggesting that partial design completion becomes more challenging when the remaining known parameters are insufficient to cover the whole design space. Interestingly, when most parameters are inpainted (over 36), the error decreases again, implying that the generation process relies on the learned global distribution rather than inconsistent local conditions. This trend highlights that intermediate levels of parameter inpainting pose the greatest difficulty, while the proposed method remains relatively robust in both low- and high-missing-parameter regimes.

## 5. Discussion
### 5.1 Zero-shot parametric engineering design generation

This study demonstrates that the proposed zero-shot sequential design generation method can achieve strong performance across multiple parametric engineering design datasets, including ship hulls, BlendedNet, and airfoil geometries. From the results shown in Sections 4.2 and 4.3, without any additional model training, the method consistently generates designs with high performance accuracy, particularly in highly structured parametric design spaces, and exhibits competitive diversity compared with diffusion-based generative models. The results indicate that the proposed approach effectively captures the relationship between performance indicators and design parameters, enabling reliable conditional design generation under varying design representations and dimensionalities. A key advantage of the proposed method lies in its zero-shot nature and its low requirement for the input data as shown in Sections 4.4 and 4.5, which eliminates the substantial computational cost and time associated with training deep generative models. This characteristic not only significantly reduces development overhead but also enables flexible and rapid deployment in practical engineering workflows, where design requirements, parameter definitions, or datasets may frequently change. Consequently, the proposed method provides a practical alternative to conventional generative models, offering a favorable balance between generation accuracy, diversity, and computational efficiency, and making it particularly suitable for real-world engineering design applications.

### 5.2 Application in geometry generation

Considering the results presented in Section 4.2, the proposed method demonstrates better performance when the random ordering is adopted during the design generation process, particularly in geometry generation tasks. Based on observations of the generation process, the authors speculate that this improvement may be attributed to the fact that the random-order generation could enable the TabPFN model to quickly perceive the rough airfoil diagram and perform detailed interpolation in subsequent generations. By observing geometrically dispersed points at early stages, the model can infer the global shape distribution of the airfoil, which subsequently guides the generation of remaining parameters in a manner analogous to interpolation rather than extrapolation. This behavior allows the model to progressively refine local geometric details while remaining anchored to a globally consistent shape. Furthermore, random ordering implicitly reduces the dependency on strict geometric adjacency during generation, alleviating error accumulation that may arise in strictly ordered boundary-following schemes.

In addition, from Table 2, the MAPE and MAE values of the airfoil with both 30 and 50 sample points are close, indicating that the random-order generation is less affected by the number of sample points. The quality of the generated geometries remains stable across different sampling resolutions, and thus varying numbers of design parameters, indicating that the method is largely insensitive to changes in geometric discretization or dimensionality. This robustness is seldom observed in the other generative model, whose performance degrades as the sampling density or parameter count increases. As a result, the proposed method is better able to maintain geometric coherence even in high-dimensional settings. These findings suggest that random-order sequential generation offers an effective strategy for geometry-based design tasks, particularly when combined with zero-shot inference and provide further evidence of the method's suitability for flexible and robust engineering design generation.

### 5.3 Practical Guidelines for Applying the Proposed Method

Based on the experimental results and observations, several practical guidelines can be provided for using the proposed sequential design generation method. The method works best for low-dimensional design problems, where it achieves good efficiency and accuracy without additional training. However, in high-dimensional cases, the inference time increases noticeably. For long-term or repeated use, training a conventional generative model, such as a conditional DDPM, is recommended to reduce overall computational cost.

The generation performance is also affected by relationships among design parameters. When strong and clear dependencies exist between parameters, better accuracy can be achieved by generating the main or global parameters first and then generating local or component-level parameters. When parameter relationships are complex or largely unknown, using a cDDPM in parallel can help avoid issues caused by the fixed generation order. For geometry-based design tasks, a random generation order is often beneficial, as it helps the model capture the overall shape early and refine details gradually, reducing error accumulation and improving robustness to different sampling resolutions and design dimensionalities

### 5.4 Limitations

Due to the limited number of reference input samples supported by TabPFN, the proposed method is unable to learn the entire data distribution in a unified manner without fine-tuning, as is commonly achieved by conventional generative models. The results reported in Section 4.3 provide clear empirical evidence of this limitation, particularly reflected in reduced coverage and sensitivity to generation strategies. Moreover, in design datasets characterized by complex interdependencies among parameters, the performance of the proposed method is influenced by the chosen parameter generation order, indicating that sequential dependency plays a non-negligible role in such settings. These factors inevitably reduce the generality of the proposed approach. Future work could focus on specific optimizations for data with strong interdependencies to broaden this method's applicability

### 6. Conclusion

This work introduces a zero-shot sequential framework based on TabPFN for parametric engineering design generation, demonstrating its effectiveness across multiple engineering datasets, including ship hulls, BlendedNet, and airfoil geometries. By leveraging TabPFN without any task-specific training or fine-tuning, the proposed method achieves competitive performance accuracy and diversity while substantially reducing computational cost and development time. Importantly, the framework requires only a limited number of reference samples for design generation, avoiding the large-scale datasets typically needed by deep generative models. Beyond full design synthesis, the proposed framework naturally supports local design generation and partial design completion by flexibly conditioning on any subset of known parameters. This low-data requirement, together with the zero-shot inference process, makes the proposed method particularly attractive for engineering scenarios where data acquisition is expensive or limited. Overall, the results indicate that zero-shot, low-sample inference can serve as a practical and flexible alternative to conventional generative models, especially in workflows that demand rapid iteration, adaptability, and localized design modification.

While the present study focuses on parametric and 2D geometry-based design tasks, several promising research directions remain open. First, extending the proposed framework to 3D geometry generation represents an important next step, where sequential generation could be applied to surface points, mesh vertices, or implicit geometric representations. Second, further investigation into high-dimensional design spaces with a larger number of parameters is warranted, particularly for complex engineering systems with strong parameter coupling and hierarchical dependencies. Finally, integrating the proposed zero-shot generation method into design optimization pipelines, such as iterative performance-driven search or human-in-the-loop optimization, could enable efficient exploration and refinement of engineering designs without repeated model retraining.

### Acknowledgement

Funding from the Natural Science and Engineering Research Council (NSERC) of Canada under the project RGPIN2019-06601 is gratefully acknowledged.

**Appendix**

**A1 Predictor Evaluation**

As mentioned in Section 4.1, we utilize the TabPFN regressor to predict the performance of the design in BlendedNet and UIUC airfoil dataset due to the lack of a calculation method provided by the dataset. In this section, we test the predictor by loading 5,000 reference performances-design data pairs to fit the model and conduct the prediction on the test set provided by the datasets. The performance of the predictor can be seen from Table 5.

Table 5: The performance of the TabPFN regressor in the prediction of BlendedNet and airfoil

|  |  | $R^2$ | MAPE (%) |
|---|---|---|---|
|  | $C_l$ | 0.994 | 3.951 |
| BlendedNet | $C_d$ | 0.993 | 3.955 |
|  | $C_{my}$ | 0.994 | 3.947 |
| airfoil 30 | $C_l/C_d$ | 0.989 | 4.935 |
| airfoil 50 | $C_l/C_d$ | 0.990 | 6.009 |

Table 5 summarizes the prediction performance across different datasets and target performance indicators. For the BlendedNet dataset, the proposed method achieves consistently high accuracy across all three output coefficients ($C_l$, $C_d$, and $C_{my}$), with $R^2$ values close to 0.994 and MAPE remaining below 4%. The results indicate that the model does not exhibit output-specific bias and is capable of stably capturing multi-output aerodynamic relationships. For the airfoil datasets, where the target performance indicator is defined as the lift-to-drag ratio ($C_l/C_d$), the method maintains a high coefficient of determination ($R^2 \approx 0.99$) for both Airfoil-30 and Airfoil-50; however, the MAPE increases to 4.94% and 6.01%, respectively. This increase can be attributed to the ratio-based performance metric, which is inherently more sensitive to prediction errors, particularly in regions with small drag values, as well as the increased nonlinearity associated with higher-dimensional design spaces. Overall, these results demonstrate that the proposed framework preserves strong global predictive fidelity as design dimensionality increases, while local relative errors become more pronounced for highly nonlinear, composite performance targets.

**A2 Strategies for Mitigating Over-Consistency in TabPFN-Based Design Generation**

As mentioned in Section 4.3, the strong consistency of the TabPFN regressor's outputs reduces the practical application value of the proposed method by limiting the diversity of generated designs. To address this limitation, two potential solutions are explored: (1) introducing slight stochastic perturbations into the generation process by adding Gaussian noise (std.: 0.0001) to each generated design parameter, and (2) varying the reference design set while keeping the generation condition unchanged. All experiments in this section are conducted using the ship hull design dataset. To evaluate the first solution, a fixed performance value is repeated 1500 times as the generation condition, and 1500 new designs are generated while adding small Gaussian noise with a standard deviation of 0.0001 to the generated design parameters. The value distributions of the first two design parameters are then analyzed and presented in Fig. 6 to examine whether diversity is introduced under identical generation conditions. In addition, the performance values of the generated designs are computed and shown in Fig. 7 to assess whether the performance characteristics are affected by noise injection. For the second solution, the generation conditions, consisting of 2000 different expected performance values, are kept unchanged, while three reference design sets are randomly sampled from the ship hull dataset. Design generation is conducted separately for each reference set, and the value distributions of the first two design parameters across the three reference sets are compared in Fig. 8.

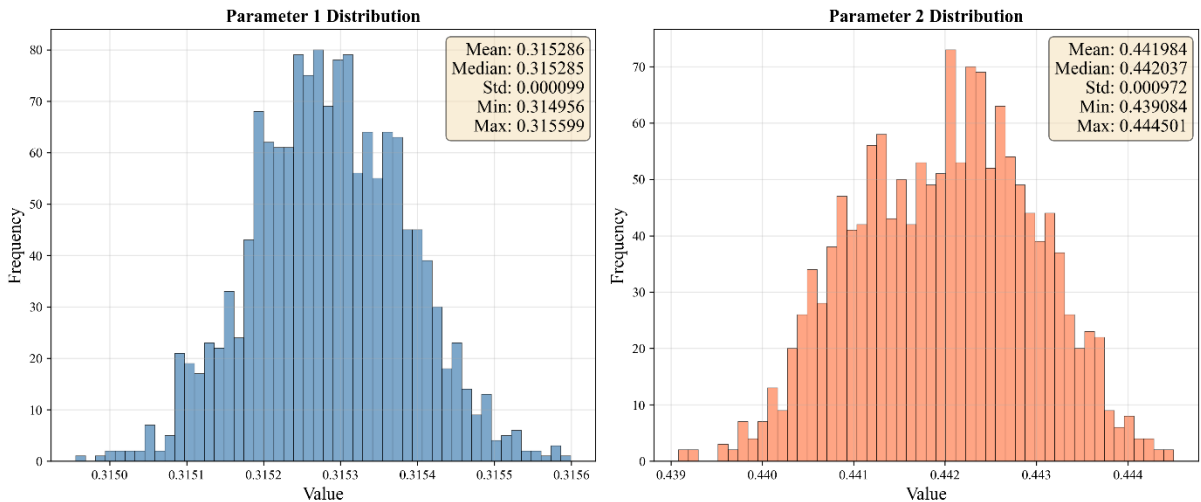

*Fig. 6 value distribution of first two generated design parameter after noise adding*

Fig. 6 shows the distributions of the first two design parameters generated under a fixed performance condition with 1500 repeated samples, where small Gaussian noise (std. = 0.0001) is added during generation. Both parameters exhibit smooth, unimodal distributions centered near their means. Parameter 1 is tightly clustered around 0.3153 (std. = 0.00099), while Parameter 2 shows a slightly broader distribution around 0.4420 (std. = 0.00097). In both cases, the close alignment between mean and median indicates that noise injections do not introduce bias but produce symmetric perturbations. Compared to the near-deterministic outputs without noise (Section 4.3), the non-zero variance confirms that controlled noise effectively increases output diversity while keeping the generated designs within a stable local neighborhood.

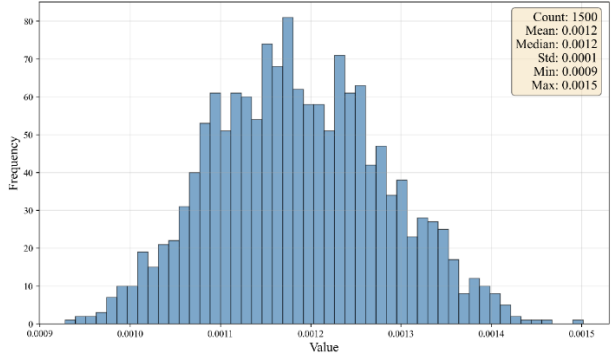

*Fig. 7 performance distribution of generated designs after noise adding*

Fig. 7 presents the performance distribution of 1500 designs generated under a fixed condition with small Gaussian noise (std. = 0.0001) added to the design parameters. The distribution is unimodal and approximately symmetric, with a mean and median of 0.0012 and a standard deviation of 0.0001, indicating stable generation within a narrow performance range. Compared to the noise-free case, performance accuracy improves, with MAPE decreasing from 13.75% to 7.63%, reported here as an empirical observation under the current setting. Together with the parameter distributions in Fig. 6, these results suggest that controlled noise injection mitigates over-consistency (Section 4.3) while preserving stable performance, with a deeper analysis of its impact left for future work.

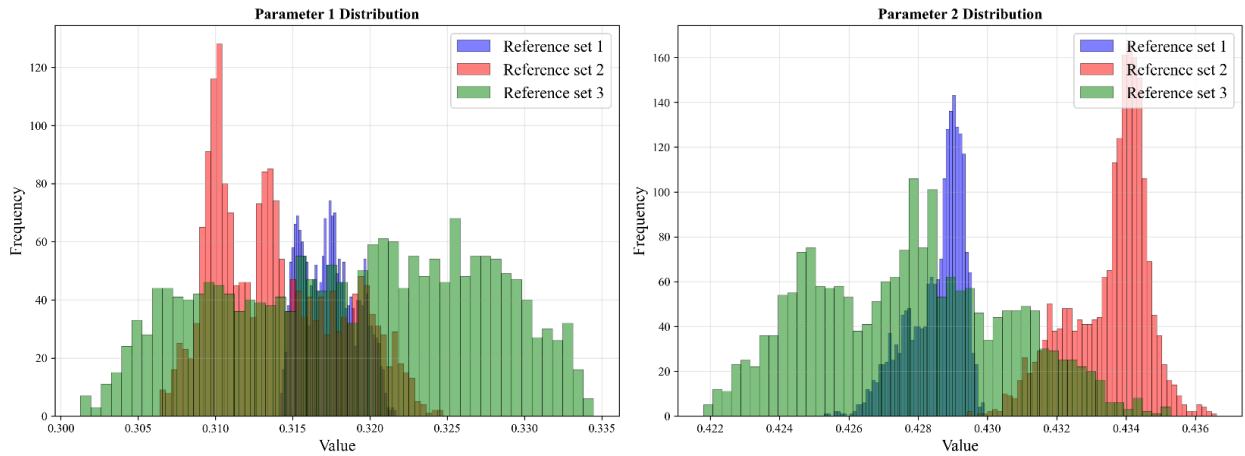

*Fig. 8 value distribution of the first two generated design parameters across three different reference design sets*

Fig. 8 compares the distributions of the first two design parameters generated using three randomly sampled reference design sets under identical generation conditions. For both parameters, the distributions show clear shifts in their central values and spread across reference sets, despite partial overlaps for Parameter 1 and well-separated peaks for Parameter 2. These results indicate that varying the reference set leads to exploration of different regions of the design space even under fixed performance constraints. Consequently, reference set selection plays a non-negligible role in shaping TabPFN-based generation and provides an effective mechanism to alleviate over-consistency without altering the generation condition. A detailed analysis of its interaction with performance constraints is left for future work.